\documentclass{article}

    \PassOptionsToPackage{numbers, compress}{natbib}

\usepackage[final]{neurips_2023}

\usepackage[utf8]{inputenc} 
\usepackage[T1]{fontenc}    
\usepackage{hyperref}       
\usepackage{url}            
\usepackage{booktabs}       
\usepackage{amsfonts}       
\usepackage{amsmath}        
\usepackage{amssymb}
\usepackage{nicefrac}       
\usepackage{microtype}      
\usepackage{xcolor}         

\usepackage{tikz}
\usepackage{pgf-pie}
\usepackage{pgfplots}
\usepgfplotslibrary{groupplots}
\pgfplotsset{compat=1.17}

\title{GECKO: Generative Language Model for English, Code and Korean}

\author{
  Sungwoo Oh \\
  KIFAI\thanks{Korea Institute of Finance and Artificial Intelligence(KIFAI) is an open community aiming to research AI technologies and share the findings to the public.} \\
  \texttt{david.oh0126@gmail.com} \\
  \And
  Donggyu Kim \\
  KIFAI \\
  \texttt{donggyukimc@gmail.com} \\
}

\begin{document}

\maketitle

\begin{abstract}
  We introduce GECKO, a bilingual large language model (LLM) optimized for Korean and English, along with programming languages. GECKO is pretrained on the balanced, high-quality corpus of Korean and English employing LLaMA architecture. In this report, we share the experiences of several efforts to build a better data pipeline for the corpus and to train our model. GECKO shows great efficiency in token generations for both Korean and English, despite its small size of vocabulary. We measure the performance on the representative benchmarks in terms of Korean, English and Code, and it exhibits great performance on KMMLU (Korean MMLU) and modest performance in English and Code, even with its smaller number of trained tokens compared to English-focused LLMs. GECKO is available to the open-source community under a permissive license. We hope our work offers a research baseline and practical insights for Korean LLM research. The model can be found at: \url{https://huggingface.co/kifai/GECKO-7B}
\end{abstract}

\section{Introduction}
Recent advances in artificial intelligence yield significant breakthroughs in the development of large language models (LLMs). Many proprietary LLMs \cite{achiam2023gpt, models2023model, team2023gemini} demonstrate human-level performances across multiple languages and on a wide range of real-world tasks \cite{chiang2023can, luo2024large, zheng2024judging}. In response to this, the open-source community has released various open large language models \cite{touvron2023llama, jiang2023mistral, jiang2024mixtral, conover2023free}, striving to match the capabilities of proprietary models.

While these open-source models have been mainly trained on English \cite{touvron2023llama, jiang2023mistral} or designed for specific use-cases such as programming \cite{li2023starcoder, wang2023codet5+, guo2024deepseek} and mathematics \cite{lewkowycz2022solving, luo2023wizardmath, yue2023mammoth}, there has been increasing demand for models proficient in other languages. This need has led to the emergence of open-source language models that show strong understanding of non-english languages such as Chinese \cite{bai2023qwen, young2024yi}, Finnish \cite{luukkonen2024poro}, and Indonesian \cite{sea_lion_2023}. They achieve impressive performance by leveraging language-specific datasets at the pretraining phase \cite{bai2023qwen, young2024yi, luukkonen2024poro, sea_lion_2023}.

Several open-source models enhance their linguistic performance by employing the following strategies: 1) language-specific continuous pretraining \cite{pipatanakul2023typhoon}, 2) vocabulary expansion \cite{l._junbum_2023}. These approaches efficiently improve the cross-lingual capabilities of the monolingual models compared to the process of pretraining models from scratch, which requires massive computational resources and extensive data.

Despite the achievements of previous Korean language models \cite{park2020koelectra, lee2020kcbert, kakaobrain2021kogpt, ko2023technical}, research on pretraining methods and applications for Korean LLM remains limited. To address this, we initiate the development of GECKO, a language model designed mainly for Korean, yet capable in English and programming languages. GECKO is pretrained from scratch, utilizing terabytes of textual data in both Korean and English to secure a strong bilingual proficiency. In the remainder of this report, we share our efforts and contributions as follows:

\begin{itemize}
    \item Data preprocessing and training methods maintaining the balance between Korean and English
    \item Demonstration of strong performance in Korean with only small amount of pretraining resources
    \item Open-sourcing our model under a permissive license to encourage further researches and applications
\end{itemize}

\section{Datasets}

\subsection{Sources}

Low-resource languages, such as Korean, have far fewer public data sources available, even if they contain data with copyright issues. In contrast, resource-rich languages like English have large, accessible data sources for training language models.

\definecolor{f1a}{HTML}{f47068}
\definecolor{f1b}{HTML}{ffb3ae}
\definecolor{f1c}{HTML}{fff4f1}
\definecolor{f1d}{HTML}{1697a6}
\definecolor{f1e}{HTML}{0e606b}
\definecolor{f1f}{HTML}{ffc24b}
\definecolor{f2a}{HTML}{d6ccc2}
\definecolor{f2b}{HTML}{e5ded8}
\definecolor{f2c}{HTML}{eeeae6}
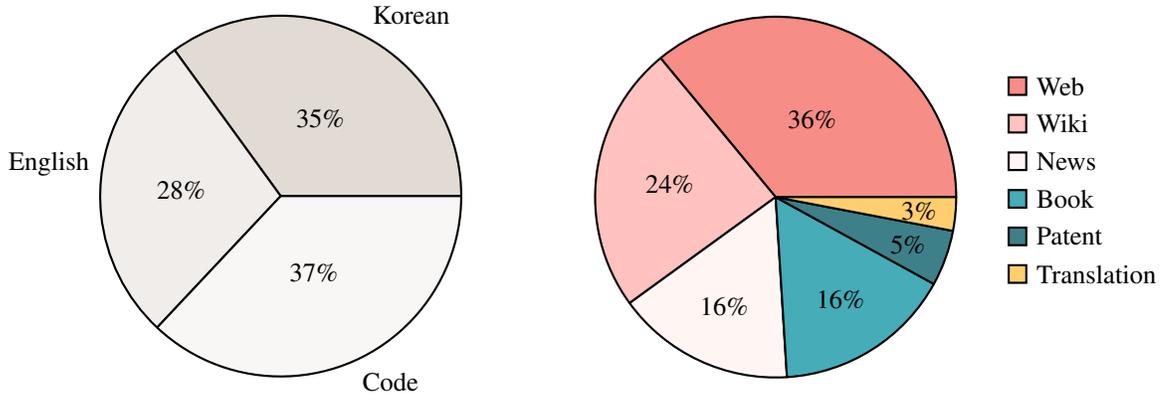
\begin{figure}[t]
    \centering
    \begin{minipage}{0.45\textwidth}
        \centering
        \begin{tikzpicture}[scale=0.8]
            \pie[color={f2a!70, f2b!55, f2c!40}]{35/Korean, 28/English, 37/Code}
        \end{tikzpicture}
    \end{minipage}%
    \hfill
    \begin{minipage}{0.45\textwidth}
        \centering
        \begin{tikzpicture}[scale=0.8]
            \pie[text=legend, color={f1a!80, f1b!80, f1c!80, f1d!80, f1e!80, f1f!80}]{36/Web, 24/Wiki, 16/News, 16/Book, 5/Patent, 3/Translation}
        \end{tikzpicture}
    \end{minipage}
    \caption{Distribution of pretraining data sources for bilingual language models. The left pie chart illustrates the proportional composition of the corpus by language, highlighting a balanced representation of 35\% Korean, 28\% English, and 37\% code to accommodate low-resource language challenges. The right pie chart details the types of data utilized, with 36\% web sources, 24\% from Wikipedia, 16\% from news articles, 16\% from books, 5\% from patents, and 3\% from translated texts. This distribution supports efforts to enhance model performance by diversifying and balancing the training data across different types and languages.}
    \label{fig:data_distribution} 
\end{figure}

\paragraph{Balancing Korean and English}
As shown in Figure \ref{fig:data_distribution}, similar to other bilingual language models \cite{bai2023qwen, luukkonen2024poro}, we aim to strike a balance between English and Korean in our pretraining corpus by down-sampling and up-sampling English and Korean corpus respectively.

\paragraph{High quality Korean corpus}
Since abundant open-source corpora for languages such as English and code already exist, and their refinement and processing significantly impact on the performance of language models \cite{penedo2023refinedweb, penedo2024fineweb}, our focus has shifted more towards improving methods of data cleansing methods. However, because high-quality Korean corpora without licensing issues are extremely limited, we collect data from Web.

\paragraph{Reasoning capability}
Additionally, research findings \cite{wei2022chain, fu2022does, huang2022towards} indicate that incorporating code data in the pretraining phase enhances the reasoning ability of language models, along with the academic perspective that treats code data as its own language. This ultimately led us to utilize three main corpora for pretraining: English, code, and Korean.

\paragraph{Language Alignment}
There is research \cite{chowdhery2023palm} on using translation datasets consisting of different language pairs for the purpose of multilingual alignment in the pretraining phase. Adopting this methodology, we train our model to align languages between English and Korean.

\begin{figure}
    \centering
    \includegraphics[width=0.8\linewidth]{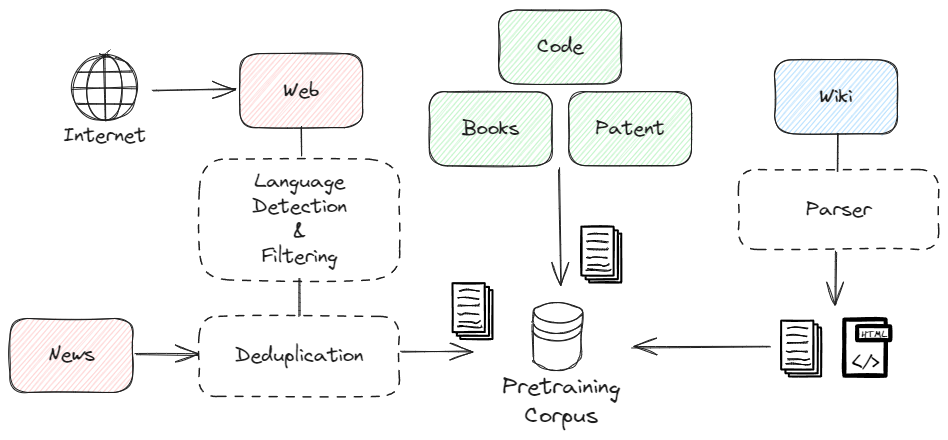}
    \caption{Pipeline for cleansing corpus}
    \label{fig:processing}
\end{figure}

\subsection{Preprocessing}
We curate and process terabytes of Korean corpus, and utilize large scale open-source corpora for English and programming languages. A sophisticated pipeline for deduplication and cleaning of raw text is implemented to obtain high-quality data as shown in Figure \ref{fig:processing}. The primary objectives of this data processing are as follows:

\begin{itemize}
    \item \textbf{Mitigate harmful content:}  Preprocess and use selective data in order to remove harmful, toxic, and biased content from the training corpus. \cite{kandpal2022deduplicating, soldaini2024dolma}.
    \item \textbf{Minimize data memorization:} The data deduplication improves robustness and generalization of the models when exposed to new, unseen data, preventing it from merely replicating patterns and generating training examples \cite{lee2021deduplicating, kassem2023preserving}.
    \item \textbf{Keep structure information:} Utilizing structural corpus including tables and lists plays a crucial role in increasing model performance.
\end{itemize}

\begin{figure}
    \centering
    \includegraphics[width=0.8\linewidth]{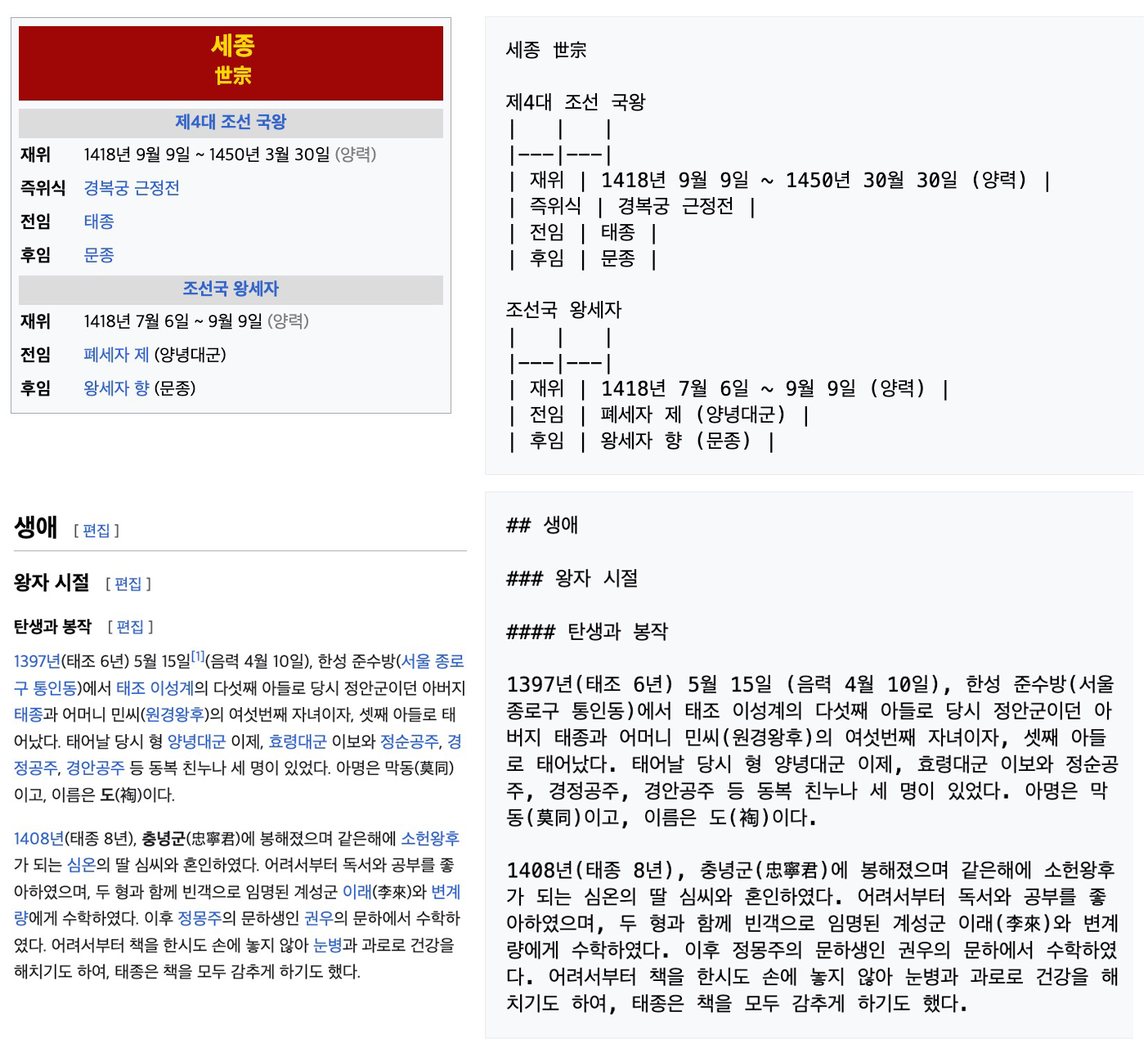}
    \caption{Example of normalization for a wiki dataset: The left image displays the original data, while the right image shows the preprocessed and normalized data in markdown format.}
    \label{fig:normalization}
\end{figure}

The training corpus includes the processing and normalization of specialized datasets such as wikis, programming code, mathematical expressions, and expert contributions. This step focuses on leveraging the structural elements inherent in these data types while carefully preserving tags and markdown features as shown in Figure \ref{fig:normalization}. These considerations allow the model to interpret and generate contextually informed and syntactically coherent outputs, significantly enhancing its utility across various applications.

\section{Pretraining}

\subsection{Tokenizer}
We train GECKO tokenizer on the balanced corpus of Korean, English, and Code. Similar to other large language models \cite{touvron2023llama, jiang2023mistral}, we utilize the Byte Pair Encoding (BPE) algorithm and train the tokenizer using Hugging Face's tokenizer. We treat all numbers as individual digits and segment unknown UTF-8 characters into bytes to avoid out-of-vocabulary issues. Additionally, we opt not to use NFKC normalization \cite{davis2001unicode}, recently reporting performance degradation on BPE-based tokenizers \cite{hoffmann2022training, rae2021scaling, le2023bloom}.

We set the total vocabulary size to 32,000, following research \cite{gowda2020finding} on optimal vocabulary size that aims to balance computational efficiency and performance considering larger vocabularies demand more computational power during the inference. We measure the efficiency of GECKO tokenizer compared to others using the following formula:

\begin{equation}
\text{Efficiency} = \left(\frac{\#\text{ of tokens}_{\text{model}}}{\#\text{ of tokens}_{\text{GECKO}}}\right) \times 100\%
\end{equation}

The metric evaluates the tokenization efficiency by comparing the total number of tokens produced by GECKO tokenizer and others. Our tokenizer demonstrates superior efficiency in processing Korean while maintaining comparable results in English and Code, contrasting to the models primarily trained in English. The result of efficiency comparison using C4 corpus \cite{raffel2020exploring} and The Stack \cite{Kocetkov2022TheStack} is illustrated in Table \ref{tab:tokenizer} and Figure \ref{fig:tokenizer}.

\begin{table}[ht]
        \centering
        \caption{Overall toeknizer efficiency with respect to GECKO.}
        \begin{tabular}{lcccccc}
            \toprule
            Tokenizer & GECKO & Polyglot-Ko & LLaMA-2 & Mistral & Gemma & GPT-4 \\
            \midrule
            Vocab. size & 32,000 & 30,080 & 32,000 & 32,000 & 256,000 & 100,277 \\
            Efficiency & 100\% & 71\% & 86\% & 92\% & 109\% & 110\% \\
            \bottomrule
        \end{tabular}
        \label{tab:tokenizer}
\end{table}

\definecolor{meta}{HTML}{1D65C1}
\definecolor{mistral}{HTML}{FF7000}
\definecolor{gpt}{HTML}{A8CD9F}
\begin{figure}[t]
    \centering
    \begin{tikzpicture}
    \begin{axis}[
        ybar,
        bar width=.5cm,
        enlarge x limits=0.25, 
        symbolic x coords={Korean, English, Code},
        xtick=data,
        ymin=0, ymax=150, 
        ylabel={Token Efficiency (\%)},
        xlabel={},
        legend style={
            at={(1.05, 0.8)},      
            anchor=north west,     
            legend columns=1,      
            draw=none,             
            row sep=7pt,           
            legend cell align=left, 
            legend image post style={xshift=1em} 
        },
        width=0.8\textwidth,
        height=8cm,
        nodes near coords,
        nodes near coords align={vertical},
        axis x line*=bottom,
        axis y line*=left
    ]
    
    \addplot[fill=black!60] coordinates {(Korean, 100) (English, 100) (Code, 100)};
    \addlegendentry{GECKO}

    \addplot[fill=gray!10] coordinates {(Korean, 101) (English, 57) (Code, 56)};
    \addlegendentry{Polyglot-Ko}

    \addplot[fill=meta!60] coordinates {(Korean, 48) (English, 111) (Code, 98)};
    \addlegendentry{LLaMA-2}
    
    \addplot[fill=mistral!60] coordinates {(Korean, 62) (English, 115) (Code, 100)};
    \addlegendentry{Mistral}
        
    \addplot[fill=gpt!60] coordinates {(Korean, 67) (English, 130) (Code, 132)};
    \addlegendentry{GPT-4}
    
    \draw [red, dashed, thick] (rel axis cs:0,0.67) -- (rel axis cs:1,0.67);
    
    \end{axis}
    \end{tikzpicture}
    \caption{Comparative analysis of tokenizer efficiency across multiple language models. This graph illustrates the performance of various tokenizers, including GECKO, Polyglot-Ko, LLaMA-2, Mistral, and GPT-4, across Korean, English, and code text corpora. The y-axis represents token efficiency as a percentage, with higher values indicating superior encoding performance relative to the tokenizer of GECKO. This analysis highlights the varying efficiency levels each model exhibits, offering insights into how effectively each tokenizer encodes multilingual and coding data. The dashed red line at 100\% serves as a benchmark for baseline efficiency.}
    \label{fig:tokenizer}
\end{figure}
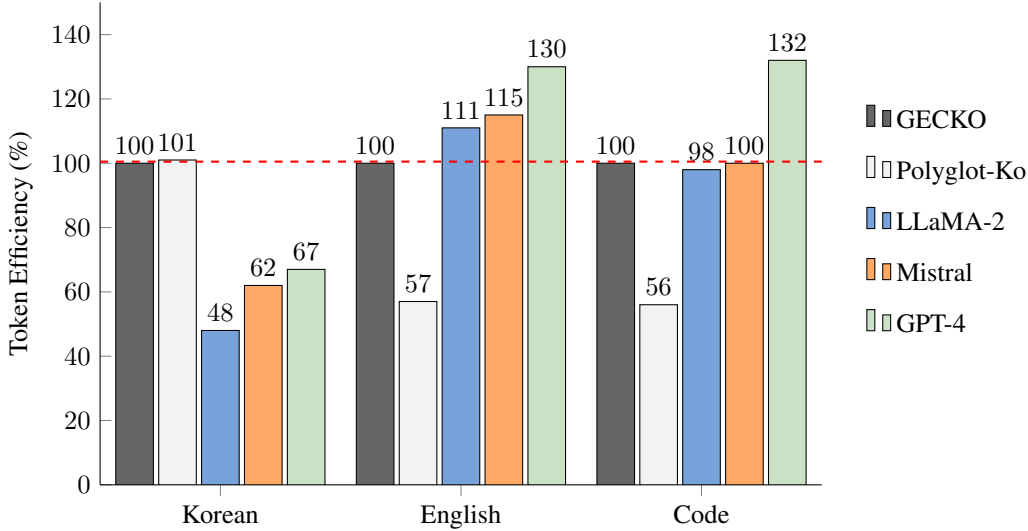

\subsection{Training Details}
GECKO adopts the classical decoder-only Transformer architecture used in LLaMA \cite{touvron2023llama}. The AdamW optimizer is employed to train the model, setting $\beta_1$ at 0.9 and $\beta_2$ at 0.95. The optimizer is configured to warm up over 10,000 iterations with a linearly increasing learning rate that peaks at 3e-4 and then decays to 3e-5 according to a cosine schedule. The model is trained with 200 billion tokens using BF16 mixed precision. Rotary positional embedding is utilized to train longer context tokens up to 8192 in length, allowing longer sequences to be understood during pretraining. We use sequence packing \cite{chung2024scaling} to assemble multiple training samples into a single sequence and use end-of-sequence token to separate the document sequences.

\subsection{Training Infrastructure}
We train our model on Google Cloud Platform and used TPUv4 with 256 chips, utilizing Fully Sharded Data Parallelism (FSDP) and model parallelism. Leveraging JAX \cite{bradbury2018jax, frostig2018compiling}, we implement the single controller programming paradigm, which enables us to manage and parallelize our training efficiently using just one Python command.

\begin{table}[h]
	\centering
        \caption{Performance evaluations across different models and benchmarks}
	\begin{tabular}{lcccc}
		\toprule
             & KMMLU & MMLU & HumanEval & MATH \\
    	Model & \textit{5-shot} & \textit{5-shot} & \textit{pass@1} & \textit{4-shot}  \\
		\midrule
            LLaMA-2 7B       & 24.2 & 45.3 & 12.8 & 2.5 \\
            Mistral 7B       & 21.0 & 62.5 & 26.2 & 12.7 \\
            Gemma 7B         & 21.1 & 64.3 & 32.3 & 24.3 \\
		  Polyglot-Ko 5.8B & 28.3 & 26.8 & 0.0 & 0.3 \\
            GECKO            & 30.7 & 28.3 & 17.7 & 4.3 \\
		\bottomrule
	\end{tabular}
	\label{tab:table}
\end{table}

\section{Evaluation}

We evaluate several pretrained open-source large language models (LLMs) released under permissive licenses. For performance assessment, we use standard academic benchmarks to evaluate knowledge and reasoning abilities \cite{hendryckstest2021}, as well as coding \cite{chen2021evaluating} and mathematics \cite{2019arXiv}. For LLaMA-2 \cite{touvron2023llama}, Mistral \cite{jiang2023mistral}, and Gemma \cite{team2024gemma}, we directly quote the scores as reported in the Gemma technical report \cite{team2024gemma}. Additionally, for the Korean evaluation set KMMLU \cite{son2024kmmlu}, we conduct our own evaluation in the same environment with previous works. The result is shown in Table \ref{tab:table}. In terms of Korean understanding (KMMLU), GECKO shows better performance compared to the evaluated models. Our model also demonstrates moderate performance in coding and mathematics.

\section{Conclusion}

GECKO is an open-source Korean pretrained LLM released under a permissive license. Our work can contribute to both academic research and the practical development of the large Korean language model pretraining. Our immediate goal is to release an improved version of the model with additional training resources. We are also preparing for instruction fine-tuning to evaluate GECKO's instruction-following ability. We believe that open-sourcing artificial intelligence technologies helps create safer products, accelerate innovation, and expand markets.

\section*{Acknowledgements}
We deeply thank the TRC Team at Google Cloud for their dedication and support, which significantly enhanced our research through provision of Cloud TPUs.

\bibliographystyle{plain}
\bibliography{references}

\begin{thebibliography}{10}

\bibitem{achiam2023gpt}
Josh Achiam, Steven Adler, Sandhini Agarwal, Lama Ahmad, Ilge Akkaya, Florencia~Leoni Aleman, Diogo Almeida, Janko Altenschmidt, Sam Altman, Shyamal Anadkat, et~al.
\newblock Gpt-4 technical report.
\newblock {\em arXiv preprint arXiv:2303.08774}, 2023.

\bibitem{bai2023qwen}
Jinze Bai, Shuai Bai, Yunfei Chu, Zeyu Cui, Kai Dang, Xiaodong Deng, Yang Fan, Wenbin Ge, Yu~Han, Fei Huang, et~al.
\newblock Qwen technical report.
\newblock {\em arXiv preprint arXiv:2309.16609}, 2023.

\bibitem{bradbury2018jax}
James Bradbury, Roy Frostig, Peter Hawkins, Matthew~James Johnson, Chris Leary, Dougal Maclaurin, George Necula, Adam Paszke, Jake VanderPlas, Skye Wanderman-Milne, et~al.
\newblock Jax: composable transformations of python+ numpy programs.
\newblock 2018.

\bibitem{chen2021evaluating}
Mark Chen, Jerry Tworek, Heewoo Jun, Qiming Yuan, Henrique Ponde de~Oliveira Pinto, Jared Kaplan, Harri Edwards, Yuri Burda, Nicholas Joseph, Greg Brockman, et~al.
\newblock Evaluating large language models trained on code.
\newblock {\em arXiv preprint arXiv:2107.03374}, 2021.

\bibitem{chiang2023can}
Cheng-Han Chiang and Hung-yi Lee.
\newblock Can large language models be an alternative to human evaluations?
\newblock {\em arXiv preprint arXiv:2305.01937}, 2023.

\bibitem{chowdhery2023palm}
Aakanksha Chowdhery, Sharan Narang, Jacob Devlin, Maarten Bosma, Gaurav Mishra, Adam Roberts, Paul Barham, Hyung~Won Chung, Charles Sutton, Sebastian Gehrmann, et~al.
\newblock Palm: Scaling language modeling with pathways.
\newblock {\em Journal of Machine Learning Research}, 24(240):1--113, 2023.

\bibitem{chung2024scaling}
Hyung~Won Chung, Le~Hou, Shayne Longpre, Barret Zoph, Yi~Tay, William Fedus, Yunxuan Li, Xuezhi Wang, Mostafa Dehghani, Siddhartha Brahma, et~al.
\newblock Scaling instruction-finetuned language models.
\newblock {\em Journal of Machine Learning Research}, 25(70):1--53, 2024.

\bibitem{conover2023free}
Mike Conover, Matt Hayes, Ankit Mathur, Jianwei Xie, Jun Wan, Sam Shah, Ali Ghodsi, Patrick Wendell, Matei Zaharia, and Reynold Xin.
\newblock Free dolly: Introducing the world’s first truly open instruction-tuned llm.
\newblock {\em Company Blog of Databricks}, 2023.

\bibitem{davis2001unicode}
Mark Davis and Martin D{\"u}rst.
\newblock Unicode normalization forms, 2001.

\bibitem{frostig2018compiling}
Roy Frostig, Matthew~James Johnson, and Chris Leary.
\newblock Compiling machine learning programs via high-level tracing.
\newblock {\em Systems for Machine Learning}, 4(9), 2018.

\bibitem{fu2022does}
Yao Fu, Hao Peng, and Tushar Khot.
\newblock How does gpt obtain its ability? tracing emergent abilities of language models to their sources.
\newblock {\em Yao Fu’s Notion}, 2022.

\bibitem{gowda2020finding}
Thamme Gowda and Jonathan May.
\newblock Finding the optimal vocabulary size for neural machine translation.
\newblock {\em arXiv preprint arXiv:2004.02334}, 2020.

\bibitem{guo2024deepseek}
Daya Guo, Qihao Zhu, Dejian Yang, Zhenda Xie, Kai Dong, Wentao Zhang, Guanting Chen, Xiao Bi, Y~Wu, YK~Li, et~al.
\newblock Deepseek-coder: When the large language model meets programming--the rise of code intelligence.
\newblock {\em arXiv preprint arXiv:2401.14196}, 2024.

\bibitem{hendryckstest2021}
Dan Hendrycks, Collin Burns, Steven Basart, Andy Zou, Mantas Mazeika, Dawn Song, and Jacob Steinhardt.
\newblock Measuring massive multitask language understanding.
\newblock {\em Proceedings of the International Conference on Learning Representations (ICLR)}, 2021.

\bibitem{hoffmann2022training}
Jordan Hoffmann, Sebastian Borgeaud, Arthur Mensch, Elena Buchatskaya, Trevor Cai, Eliza Rutherford, Diego de~Las Casas, Lisa~Anne Hendricks, Johannes Welbl, Aidan Clark, et~al.
\newblock Training compute-optimal large language models.
\newblock {\em arXiv preprint arXiv:2203.15556}, 2022.

\bibitem{huang2022towards}
Jie Huang and Kevin Chen-Chuan Chang.
\newblock Towards reasoning in large language models: A survey.
\newblock {\em arXiv preprint arXiv:2212.10403}, 2022.

\bibitem{jiang2023mistral}
Albert~Q Jiang, Alexandre Sablayrolles, Arthur Mensch, Chris Bamford, Devendra~Singh Chaplot, Diego de~las Casas, Florian Bressand, Gianna Lengyel, Guillaume Lample, Lucile Saulnier, et~al.
\newblock Mistral 7b.
\newblock {\em arXiv preprint arXiv:2310.06825}, 2023.

\bibitem{jiang2024mixtral}
Albert~Q Jiang, Alexandre Sablayrolles, Antoine Roux, Arthur Mensch, Blanche Savary, Chris Bamford, Devendra~Singh Chaplot, Diego de~las Casas, Emma~Bou Hanna, Florian Bressand, et~al.
\newblock Mixtral of experts.
\newblock {\em arXiv preprint arXiv:2401.04088}, 2024.

\bibitem{kandpal2022deduplicating}
Nikhil Kandpal, Eric Wallace, and Colin Raffel.
\newblock Deduplicating training data mitigates privacy risks in language models.
\newblock In {\em International Conference on Machine Learning}, pages 10697--10707. PMLR, 2022.

\bibitem{kassem2023preserving}
Aly Kassem, Omar Mahmoud, and Sherif Saad.
\newblock Preserving privacy through dememorization: An unlearning technique for mitigating memorization risks in language models.
\newblock In {\em Proceedings of the 2023 Conference on Empirical Methods in Natural Language Processing}, pages 4360--4379, 2023.

\bibitem{kakaobrain2021kogpt}
Ildoo Kim, Gunsoo Han, Jiyeon Ham, and Woonhyuk Baek.
\newblock Kogpt: Kakaobrain korean(hangul) generative pre-trained transformer.
\newblock \url{https://github.com/kakaobrain/kogpt}, 2021.

\bibitem{ko2023technical}
Hyunwoong Ko, Kichang Yang, Minho Ryu, Taekyoon Choi, Seungmu Yang, Sungho Park, et~al.
\newblock A technical report for polyglot-ko: Open-source large-scale korean language models.
\newblock {\em arXiv preprint arXiv:2306.02254}, 2023.

\bibitem{Kocetkov2022TheStack}
Denis Kocetkov, Raymond Li, Loubna Ben~Allal, Jia Li, Chenghao Mou, Carlos Muñoz~Ferrandis, Yacine Jernite, Margaret Mitchell, Sean Hughes, Thomas Wolf, Dzmitry Bahdanau, Leandro von Werra, and Harm de~Vries.
\newblock The stack: 3 tb of permissively licensed source code.
\newblock {\em Preprint}, 2022.

\bibitem{l._junbum_2023}
{L. Junbum}.
\newblock llama-2-ko-7b (revision 4a9993e), 2023.

\bibitem{le2023bloom}
Teven Le~Scao, Angela Fan, Christopher Akiki, Ellie Pavlick, Suzana Ili{\'c}, Daniel Hesslow, Roman Castagn{\'e}, Alexandra~Sasha Luccioni, Fran{\c{c}}ois Yvon, Matthias Gall{\'e}, et~al.
\newblock Bloom: A 176b-parameter open-access multilingual language model.
\newblock 2023.

\bibitem{lee2020kcbert}
Junbum Lee.
\newblock Kcbert: Korean comments bert.
\newblock In {\em Annual Conference on Human and Language Technology}, pages 437--440. Human and Language Technology, 2020.

\bibitem{lee2021deduplicating}
Katherine Lee, Daphne Ippolito, Andrew Nystrom, Chiyuan Zhang, Douglas Eck, Chris Callison-Burch, and Nicholas Carlini.
\newblock Deduplicating training data makes language models better.
\newblock {\em arXiv preprint arXiv:2107.06499}, 2021.

\bibitem{lewkowycz2022solving}
Aitor Lewkowycz, Anders Andreassen, David Dohan, Ethan Dyer, Henryk Michalewski, Vinay Ramasesh, Ambrose Slone, Cem Anil, Imanol Schlag, Theo Gutman-Solo, et~al.
\newblock Solving quantitative reasoning problems with language models.
\newblock {\em Advances in Neural Information Processing Systems}, 35:3843--3857, 2022.

\bibitem{li2023starcoder}
Raymond Li, Loubna~Ben Allal, Yangtian Zi, Niklas Muennighoff, Denis Kocetkov, Chenghao Mou, Marc Marone, Christopher Akiki, Jia Li, Jenny Chim, et~al.
\newblock Starcoder: may the source be with you!
\newblock {\em arXiv preprint arXiv:2305.06161}, 2023.

\bibitem{luo2023wizardmath}
Haipeng Luo, Qingfeng Sun, Can Xu, Pu~Zhao, Jianguang Lou, Chongyang Tao, Xiubo Geng, Qingwei Lin, Shifeng Chen, and Dongmei Zhang.
\newblock Wizardmath: Empowering mathematical reasoning for large language models via reinforced evol-instruct.
\newblock {\em arXiv preprint arXiv:2308.09583}, 2023.

\bibitem{luo2024large}
Xiaoliang Luo, Akilles Rechardt, Guangzhi Sun, Kevin~K Nejad, Felipe Y{\'a}{\~n}ez, Bati Yilmaz, Kangjoo Lee, Alexandra~O Cohen, Valentina Borghesani, Anton Pashkov, et~al.
\newblock Large language models surpass human experts in predicting neuroscience results.
\newblock {\em arXiv preprint arXiv:2403.03230}, 2024.

\bibitem{luukkonen2024poro}
Risto Luukkonen, Jonathan Burdge, Elaine Zosa, Aarne Talman, Ville Komulainen, V{\"a}in{\"o} Hatanp{\"a}{\"a}, Peter Sarlin, and Sampo Pyysalo.
\newblock Poro 34b and the blessing of multilinguality.
\newblock {\em arXiv preprint arXiv:2404.01856}, 2024.

\bibitem{models2023model}
Claude Models.
\newblock Model card and evaluations for claude models, 2023.

\bibitem{park2020koelectra}
Jangwon Park.
\newblock Koelectra: Pretrained electra model for korean.
\newblock \url{https://github.com/monologg/KoELECTRA}, 2020.

\bibitem{penedo2024fineweb}
Guilherme Penedo, Hynek Kydlíček, Leandro von Werra, and Thomas Wolf.
\newblock Fineweb, 2024.

\bibitem{penedo2023refinedweb}
Guilherme Penedo, Quentin Malartic, Daniel Hesslow, Ruxandra Cojocaru, Alessandro Cappelli, Hamza Alobeidli, Baptiste Pannier, Ebtesam Almazrouei, and Julien Launay.
\newblock The refinedweb dataset for falcon llm: outperforming curated corpora with web data, and web data only.
\newblock {\em arXiv preprint arXiv:2306.01116}, 2023.

\bibitem{pipatanakul2023typhoon}
Kunat Pipatanakul, Phatrasek Jirabovonvisut, Potsawee Manakul, Sittipong Sripaisarnmongkol, Ruangsak Patomwong, Pathomporn Chokchainant, and Kasima Tharnpipitchai.
\newblock Typhoon: Thai large language models.
\newblock {\em arXiv preprint arXiv:2312.13951}, 2023.

\bibitem{rae2021scaling}
Jack~W Rae, Sebastian Borgeaud, Trevor Cai, Katie Millican, Jordan Hoffmann, Francis Song, John Aslanides, Sarah Henderson, Roman Ring, Susannah Young, et~al.
\newblock Scaling language models: Methods, analysis \& insights from training gopher.
\newblock {\em arXiv preprint arXiv:2112.11446}, 2021.

\bibitem{raffel2020exploring}
Colin Raffel, Noam Shazeer, Adam Roberts, Katherine Lee, Sharan Narang, Michael Matena, Yanqi Zhou, Wei Li, and Peter~J Liu.
\newblock Exploring the limits of transfer learning with a unified text-to-text transformer.
\newblock {\em Journal of machine learning research}, 21(140):1--67, 2020.

\bibitem{2019arXiv}
Hill Saxton, Grefenstette and Kohli.
\newblock Analysing mathematical reasoning abilities of neural models.
\newblock {\em arXiv:1904.01557}, 2019.

\bibitem{sea_lion_2023}
AI~Singapore.
\newblock Sea-lion (southeast asian languages in one network): A family of large language models for southeast asia.
\newblock \url{https://github.com/aisingapore/sealion}, 2023.

\bibitem{soldaini2024dolma}
Luca Soldaini, Rodney Kinney, Akshita Bhagia, Dustin Schwenk, David Atkinson, Russell Authur, Ben Bogin, Khyathi Chandu, Jennifer Dumas, Yanai Elazar, et~al.
\newblock Dolma: An open corpus of three trillion tokens for language model pretraining research.
\newblock {\em arXiv preprint arXiv:2402.00159}, 2024.

\bibitem{son2024kmmlu}
Guijin Son, Hanwool Lee, Sungdong Kim, Seungone Kim, Niklas Muennighoff, Taekyoon Choi, Cheonbok Park, Kang~Min Yoo, and Stella Biderman.
\newblock Kmmlu: Measuring massive multitask language understanding in korean.
\newblock {\em arXiv preprint arXiv:2402.11548}, 2024.

\bibitem{team2023gemini}
Gemini Team, Rohan Anil, Sebastian Borgeaud, Yonghui Wu, Jean-Baptiste Alayrac, Jiahui Yu, Radu Soricut, Johan Schalkwyk, Andrew~M Dai, Anja Hauth, et~al.
\newblock Gemini: a family of highly capable multimodal models.
\newblock {\em arXiv preprint arXiv:2312.11805}, 2023.

\bibitem{team2024gemma}
Gemma Team, Thomas Mesnard, Cassidy Hardin, Robert Dadashi, Surya Bhupatiraju, Shreya Pathak, Laurent Sifre, Morgane Rivi{\`e}re, Mihir~Sanjay Kale, Juliette Love, et~al.
\newblock Gemma: Open models based on gemini research and technology.
\newblock {\em arXiv preprint arXiv:2403.08295}, 2024.

\bibitem{touvron2023llama}
Hugo Touvron, Louis Martin, Kevin Stone, Peter Albert, Amjad Almahairi, Yasmine Babaei, Nikolay Bashlykov, Soumya Batra, Prajjwal Bhargava, Shruti Bhosale, et~al.
\newblock Llama 2: Open foundation and fine-tuned chat models.
\newblock {\em arXiv preprint arXiv:2307.09288}, 2023.

\bibitem{wang2023codet5+}
Yue Wang, Hung Le, Akhilesh~Deepak Gotmare, Nghi~DQ Bui, Junnan Li, and Steven~CH Hoi.
\newblock Codet5+: Open code large language models for code understanding and generation.
\newblock {\em arXiv preprint arXiv:2305.07922}, 2023.

\bibitem{wei2022chain}
Jason Wei, Xuezhi Wang, Dale Schuurmans, Maarten Bosma, Fei Xia, Ed~Chi, Quoc~V Le, Denny Zhou, et~al.
\newblock Chain-of-thought prompting elicits reasoning in large language models.
\newblock {\em Advances in neural information processing systems}, 35:24824--24837, 2022.

\bibitem{young2024yi}
Alex Young, Bei Chen, Chao Li, Chengen Huang, Ge~Zhang, Guanwei Zhang, Heng Li, Jiangcheng Zhu, Jianqun Chen, Jing Chang, et~al.
\newblock Yi: Open foundation models by 01. ai.
\newblock {\em arXiv preprint arXiv:2403.04652}, 2024.

\bibitem{yue2023mammoth}
Xiang Yue, Xingwei Qu, Ge~Zhang, Yao Fu, Wenhao Huang, Huan Sun, Yu~Su, and Wenhu Chen.
\newblock Mammoth: Building math generalist models through hybrid instruction tuning.
\newblock {\em arXiv preprint arXiv:2309.05653}, 2023.

\bibitem{zheng2024judging}
Lianmin Zheng, Wei-Lin Chiang, Ying Sheng, Siyuan Zhuang, Zhanghao Wu, Yonghao Zhuang, Zi~Lin, Zhuohan Li, Dacheng Li, Eric Xing, et~al.
\newblock Judging llm-as-a-judge with mt-bench and chatbot arena.
\newblock {\em Advances in Neural Information Processing Systems}, 36, 2024.

\end{thebibliography}

\end{document}